\documentclass[10pt, conference, compsocconf]{IEEEtran}
\ifCLASSINFOpdf
\usepackage[pdftex]{graphicx}
\graphicspath{{../pdf/}{../jpeg/}}
\DeclareGraphicsExtensions{.pdf,.jpeg,.png}
\else
\fi

\usepackage[hidelinks]{hyperref}
\hypersetup{breaklinks=true}

\usepackage{multirow}
\usepackage{color}

\usepackage{amsmath}
\usepackage{amssymb}
\usepackage{pifont}
\newcommand{\cmark}{\ding{51}}%
\newcommand{\xmark}{\ding{55}}%

\begin{document}
	\title{$M^3$T: Multi-Modal Continuous Valence-Arousal Estimation in the Wild}
	
	\author{\IEEEauthorblockN{
			Yuan-Hang Zhang\IEEEauthorrefmark{1}\IEEEauthorrefmark{2},
			Rulin Huang\IEEEauthorrefmark{1}\IEEEauthorrefmark{2},
			Jiabei Zeng\IEEEauthorrefmark{1}, 
			Shiguang Shan\IEEEauthorrefmark{1}\IEEEauthorrefmark{2} and 
			Xilin Chen\IEEEauthorrefmark{1}\IEEEauthorrefmark{2}
		}
		\IEEEauthorblockA{\IEEEauthorrefmark{1}Key Lab of Intelligent Information Processing, Chinese Academy of Sciences (CAS)\\
			Institute of Computing Technology, CAS, 
			Beijing 100190, China\\ \{jiabei.zeng, sgshan, xlchen\}@ict.ac.cn}
		\IEEEauthorblockA{\IEEEauthorrefmark{2}School of Computer Science and Technology\\
			University of Chinese Academy of Sciences, 
			Beijing 100049, China\\ \{zhangyuanhang15, huangrulin18\}@mails.ucas.ac.cn}
	}

	\maketitle
	
	\begin{abstract}
This report describes a multi-modal multi-task ($M^3$T) approach underlying our submission to the valence-arousal estimation track of the Affective Behavior Analysis in-the-wild (ABAW) Challenge, held in conjunction with the IEEE International Conference on Automatic Face and Gesture Recognition (FG) 2020. 
In the proposed $M^3$T framework, we fuse both visual features from videos and acoustic features from the audio tracks to estimate the valence and arousal. 
The spatio-temporal visual features are extracted with a 3D convolutional network and a bidirectional recurrent neural network.
Considering the correlations between valence / arousal, emotions, and facial actions, we also explores mechanisms to benefit from other tasks.
We evaluated the $M^3$T framework on the validation set provided by ABAW and it significantly outperforms the baseline method.

	\end{abstract}

	\IEEEpeerreviewmaketitle
	
\section{Introduction}
Automatically understanding human affect is of great importance in human-machine interactions. 
Psychologists have developed the circumplex model of emotion~\cite{russell1980circumplex} to describe peoples' state of feeling. 
In the circumplex model of emotion, valence (i.e., how positive or negative an emotion is) and arousal (i.e., how powerful an emotion is) are the two dimensions that can be linked to affective and cognitive response. 
Researchers in computer science have made great efforts in estimating the affective states (i.e. valence and arousal) using visual or audio signals~\cite{DBLP:journals/corr/abs-1910-04855, DBLP:journals/ijcv/KolliasTNPZSKZ19, DBLP:journals/corr/abs-1811-07771,DBLP:conf/cvpr/KolliasNKZZ17}. 

However, valence and arousal are not the only way to represent human affect. There are two other widely adopted approaches: through categorical emotions~\cite{ekman1992argument} (e.g., happiness, sadness, anger, fear, disgust, surprise, etc.), and through facial actions (e.g., facial action units defined by the Facial Action Coding System~\cite{friesen1978facial}). 
The connections between the affective dimensions and the two other approaches are often ignored. 
Recently, Kollias et al.~\cite{DBLP:journals/corr/abs-1910-04855,DBLP:journals/corr/abs-1811-07770} collected a large scale in-the-wild dataset, Aff-Wild2, which is not only annotated with valence and arousal, but also with categorical emotions and eight facial action units. 
Based upon the newly collected benchmark, we propose a multi-modal multi-task ($M^3$T) framework to estimate the continuous valence and arousal, where valence-arousal estimation benefits from the emotion recognition task.

\begin{figure}[!t]
	\centering
	\includegraphics[width=\columnwidth]{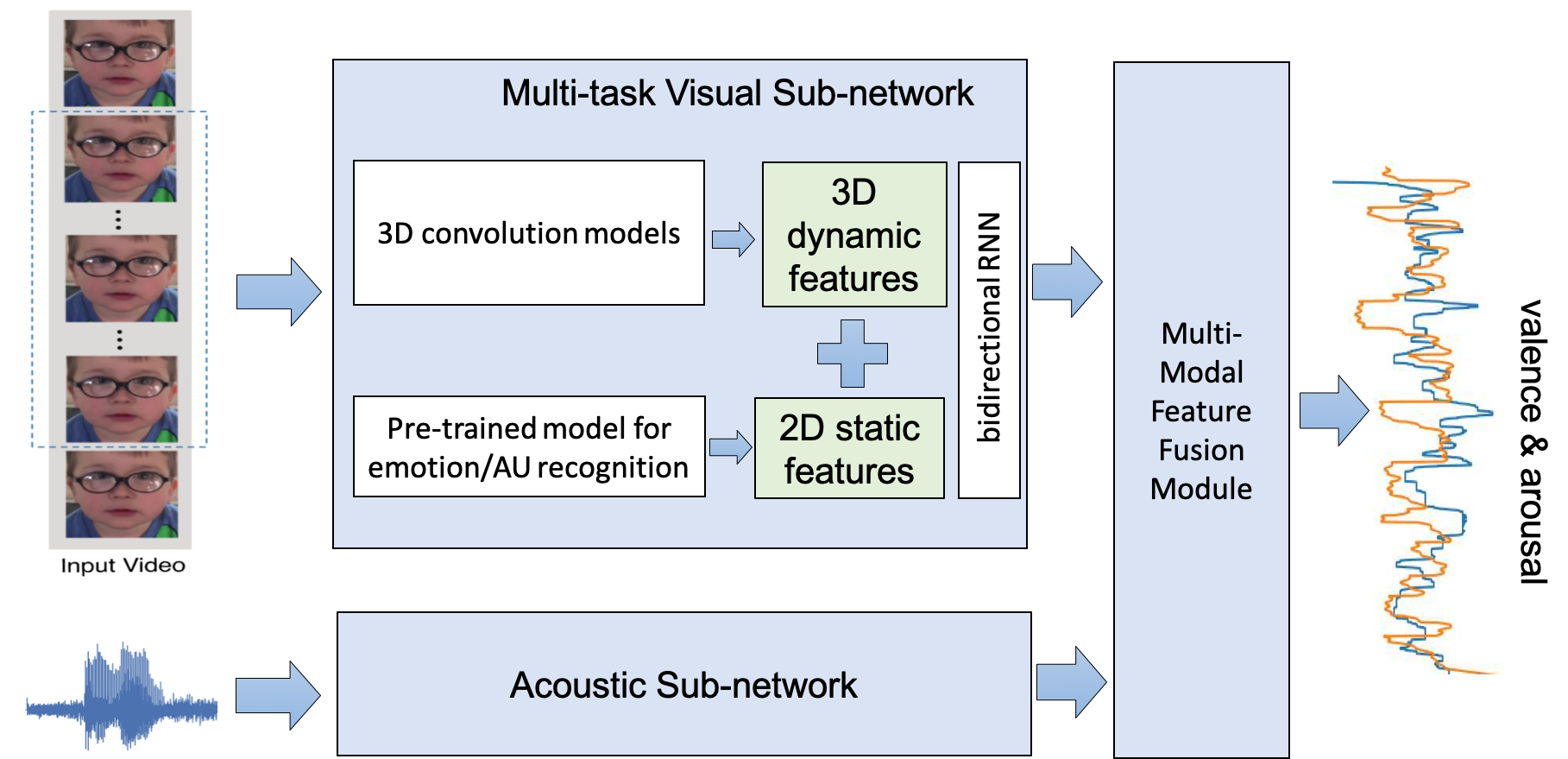}
	\caption{Overview of the proposed multi-modal multi-task ($M^3$T) framework.}
		\label{fig:overview}
\end{figure}

Fig.~\ref{fig:overview} illustrates the main idea of the proposed $M^3$T framework. Given the videos and their corresponding audio tracks, $M^3$T first extracts the visual features through a multi-task visual subnetwork, and extracts the audio features with an acoustic sub-network. 
In the multi-task visual subnetwork, we explore two mechanisms to benefit from the other tasks: training with losses for several tasks, and concatenating features from different tasks.
Then, a late-fusion mechanism is used to fuse the two features. 
The ultimate features are used to estimate valence and arousal.
	
\section{The Multi-Modal Multi-Task ($M^3$T) Framework}
The proposed $M^3$T framework consists of three parts: multi-task visual network, acoustic network, and the multi-modal feature fusion module. 
In this section, we provide details of the three components.
	
	\begin{figure*}[!t]
	\centering
	\includegraphics[width=0.85\textwidth]{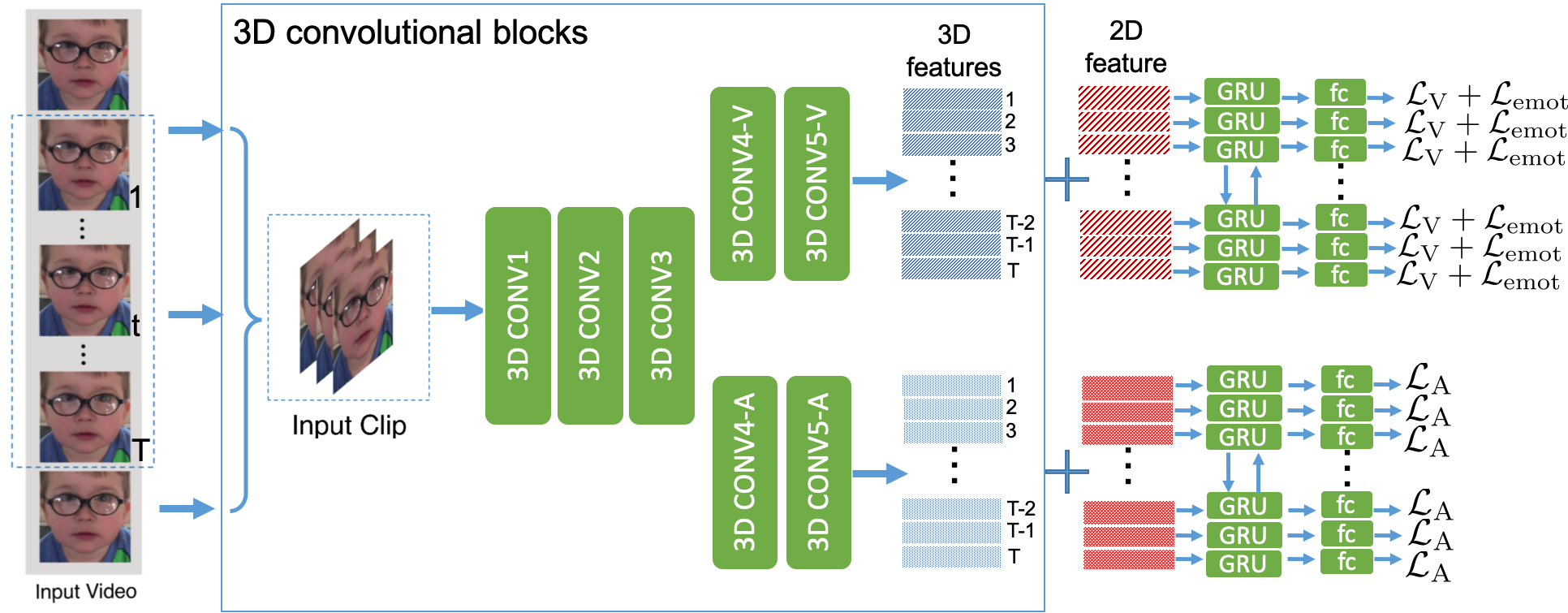}
	\caption{\textbf{Architecture of the multi-task visual sub-network.} Given the input clip, 3D features for arousal and valence are extracted from two-branch 3D convolutional blocks. Then, the 3D features are concatenated with 2D features pre-trained for other tasks. The concatenated features for each frame are encoded by bi-GRUs and passed to fully-connected layers for final predictions.}
	\label{fig:framework}
	\end{figure*}
	
\subsection{Multi-Task Visual Network}
We used categorical emotion recognition and AU detection to assist the valence estimation, because arousal can be reflected by the intensity of facial actions, while valence estimation is highly related to categorical emotion recognition.
Valence indicates how pleasant or unpleasant a person is.
It is intuitive to rate a ``happy" emotion with a high valence score, and to rate a ``sad" emotion with a low valence score. 
Due to the differences between valence and arousal, the multi-task visual network have two branches for the two affective dimensions, respectively.

Fig.~\ref{fig:framework} shows the details of the multi-task visual network. 
The multi-task visual network follows the V2P architecture~\cite{DBLP:journals/corr/abs-1807-05162}, which has been successfully applied to visual speech recognition. 
Under this architecture, we first extract spatio-temporal features with 3D convolutional blocks from a given video clip, and then aggregate these features through bidirectional recurrent neural networks. 

The multi-task visual network leverages the information from categorical emotions and facial actions with two mechanisms. 
First, we consider both the 3D features from the 3D convolution blocks and the 2D features a 2D network. 
The 2D static features are from extracted from a pretrained emotion recognition model and an AU detection model.
Second, the architecture are trained with losses for multiple tasks: $\mathcal{L}_{V}$ for valence estimation,  $\mathcal{L}_{A}$ for arousal estimation, and the cross-entropy loss $\mathcal{L}_{emot}$ for emotion recognition. 
Below, we present details of the 3D convolutional blocks, 2D static features, recurrent layers, and the losses.

{\bf 3D convolutional blocks:}
As shown in Fig.~\ref{fig:framework}, given $T$ input frames, we used a 3D VGG-like backbone to extract the spatial-temporal features for every frame.
Considering the difference between valence and arousal, the overall visual network has two branches, i.e., valence branch and arousal branch.
In the 3D convolutional blocks, the two branches share the first three convolutional layers.
Then, each branch has two specific convolutional layers. 
Therefore, through the 3D convolutional blocks, we obtain 3D features for valence and arousal, respectively.
To initialize the network properly, we first pretrain the model for video-based face recognition on $1,000$ selected identities in the development set of the VoxCeleb2 dataset~\cite{DBLP:conf/interspeech/ChungNZ18}, which consists of talking faces recorded under a variety of in-the-wild conditions.

{\bf 2D static features:} 
We extract emotion and AU features for each individual frame as 2D static features. 
Considering their different correlations with valence and arousal, we initially concatenate emotion features with the 3D valence features, and AU features with the 3D arousal features. It is also possible to concatenate emotion features with the arousal features; we discuss the impact of this choice in Sec.~\ref{ssec:results}.

For emotion features, we use $512$-dimensional features from the average-pooling layer of an SENet-101~\cite{DBLP:conf/cvpr/HuSS18} model pretrained on a large-scale dataset for facial expression recognition, which is a union of AffectNet~\cite{DBLP:journals/taffco/MollahosseiniHM19}, RAF-DB~\cite{DBLP:conf/cvpr/0001DD17}, and $300,000$ privately collected images. 
For AU features, we use the self-supervised $256$-dimensional encoder features using TCAE~\cite{DBLP:conf/cvpr/LiZSC19}, which is trained on the union of VoxCeleb1~\cite{DBLP:conf/interspeech/NagraniCZ17} and VoxCeleb2~\cite{DBLP:conf/interspeech/ChungNZ18} datasets. 
As these models are computationally expensive and trained at a different resolution ($256\times 256$), we dump frozen features to disk, and retrieve pre-computed features on-the-fly during training.

{\bf Recurrent layers:} Then, for each time step $t$, the corresponding 3D features and 2D features are concatenated along the channel dimension, and fed to a $2$-layer, $1024$-cell bidirectional recurrent neural network with Gated Recurrent Units (GRU) before two fully-connected layers.
After the fully-connected layers, we obtain the expected outputs, i.e., the estimated valence and arousal.

{\bf Loss $\mathcal{L}_\mathrm{V}$, $\mathcal{L}_\mathrm{A}$ for valence and arousal estimation:} 
We aim to maximize the agreement between the annotations and predictions by maximizing their Canonical Concordance Coefficients (CCC). 
Therefore, we minimize the formulated losses $\mathcal{L}_\mathrm{V}$ and $\mathcal{L}_\mathrm{A}$, respectively:
\begin{align}
\mathcal{L}_{\mathrm{V}} &= 1 - \rho_v, \\
\mathcal{L}_{\mathrm{A}} &= 1 - \rho_a,
\end{align}
where $\rho_v$ (resp.~$\rho_a$) are the CCCs between ground truth valence (resp.~arousal) values and the predicted valence (resp.~arousal) values. 
For an explanation of CCCs, please refer to Eq.~\eqref{eq:ccc} in Sec.~\ref{ssec:impl-details}.

{\bf Loss $\mathcal{L}_\mathrm{emot}$ for emotion recognition:} 
This is the standard categorical cross-entropy loss for classification:
\begin{equation}
\mathcal{L}_\mathrm{emot} =  \sum_{i=1}^7 p_i \log {\hat{p}_i},
\end{equation}
where $p_i\in {0,1}$ is ground truth, and $\hat{p}_i \in [0,1]$ is the prediction for each of the seven expression classes.
Note that we only apply this loss to frames that are annotated with expression.

The final loss is a weighted average of the two losses:
\begin{equation}
	\mathcal{L} = 0.5( \mathcal{L}_\mathrm{V} +\mathcal{L}_\mathrm{A} ) + \lambda \mathcal{L}_\mathrm{emot},
\end{equation}
where $\lambda=0.5$ is a balancing factor used to stabilize training.
We do not further exploit AU presence labels, as only $23$ of the videos have such annotations, and adding this information did not improve performance in our preliminary experiments.

\subsection{Acoustic Network}
The input acoustic features are stacked log-Mel spectrogram energies, synchronized with the video to yield one $200$-dimensional feature per time step. The features within each window are encoded with a $2$-layer, $512$-cell bidirectional GRU, and passed through an MLP with two fully-connected layers which transform the features as $512\to 512\to 512$. 

\subsection{Multi-Modal Feature Fusion Module}
To train the audio-visual joint model, we remove the fully-connected layers from the single-modality models. The output $2048$-dimensional visual features is first projected to $512$ dimensions with a fully connected layer, and concatenated with the $512$-dimensional GRU outputs from the audio subnetwork, in a late-fusion fashion. We finally pass the concatenated features through a two-layer bidirectional GRU and two fully-connected layers to obtain the final predictions.

We note that visual and acoustic information are not equally informative at each time step; for example, the person may be acting in silence, or temporarily invisible. Inspired by this observation, instead of simple feature concatenation, we propose another fusion scheme, which aggregates information from the two modalities using an attention mechanism.

Formally, denote the visual features at time $t$ by $v_t$ and acoustic features by $a_t$. We implement two scoring functions, $h_t^v$ and $h_t^a$ which derive ``quality scores" for the visual and audio modality based on the features. In our experiments, we instantiate $h$ as a one-layer bi-GRU with $128$ hidden units, followed by the sigmoid function. Finally we compute a fused representation $f_t$ for each time step, which is then passed to the GRU and fully-connected layers for final predictions:
	\begin{align}
	\alpha_t^{\mathrm m} &= \frac{\exp(h_t^{\mathrm m})}{\sum_{\mathrm m\in\{v,a\}}\exp(h_t^{\mathrm m})},\\
	f_t &= \alpha_t^v v_t + \alpha_t^a a_t.
	\end{align}
	
	\begin{figure}[!t]
		\centering
		\includegraphics[width=\columnwidth]{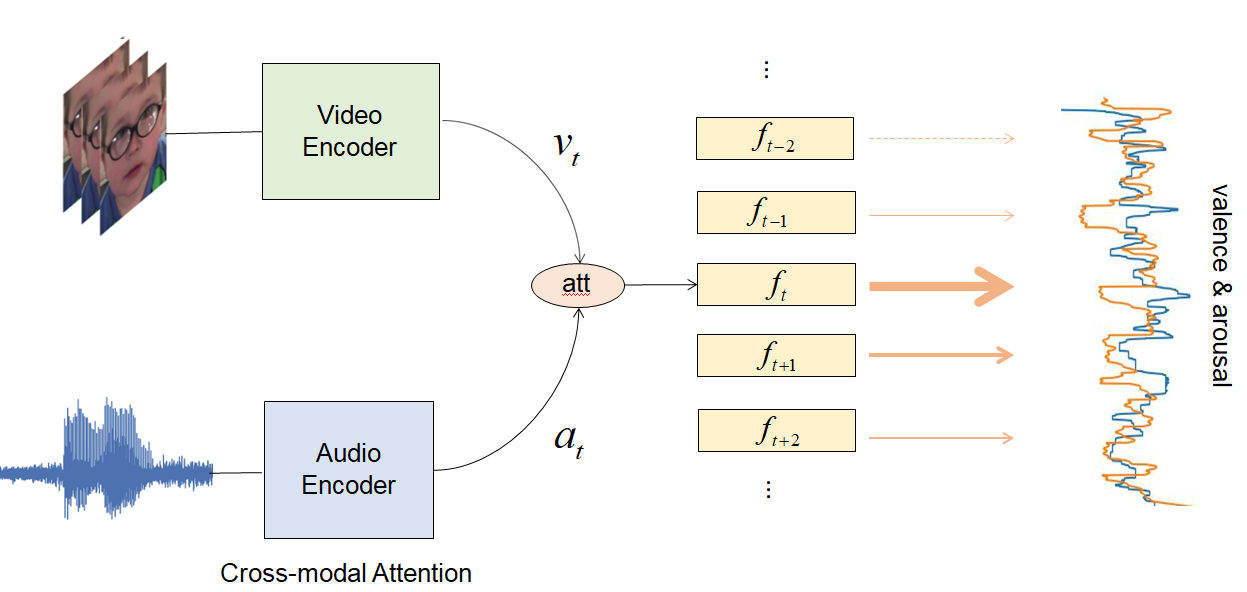}
		\caption{\textbf{An attentional multi-modal feature fusion scheme.} At each time step, we re-weight audio and visual information with an attention mechanism, and aggregate the features for final prediction.}
		\label{fig:attention}
	\end{figure}
	\section{Experiments}
	\subsection{Implementation Details}\label{ssec:impl-details}
	\textbf{Dataset.} We use the Aff-Wild2 dataset~\cite{DBLP:journals/corr/abs-1811-07770,DBLP:journals/corr/abs-1910-04855}, which contains $545$ videos with annotations for valence-arousal, facial expression and facial action units. It is an extension of the previous Aff-Wild dataset~\cite{DBLP:conf/cvpr/ZafeiriouKNPZK17}, and currently the largest audiovisual in-the-wild database annotated for valence and arousal. According to the partition and annotations provided by the ABAW 2020 challenge organizers, there are $351$, $71$, and $139$ subjects in the training, validation and test subsets respectively for the VA estimation track.
	
	\textbf{Video preprocessing.} We first run face detection on the provided videos using the RetinaFace detector~\cite{DBLP:journals/corr/abs-1905-00641} with the ResNet-50 backbone. 
	The detected faces are grouped with an IoU-based tracker, aligned to a canonical template using the five detected landmarks with a similarity transformation, and cropped to $128\times 128$. 
	Additionally, we smooth the facial landmarks with a temporal Gaussian kernel, but only if the variance of the bounding box coordinates is below a conservative threshold (since the subjects sometimes move dramatically, e.g. jumping for joy, lying down on bed, riding a roller coaster).  
	A vector of zeros is used in lieu of the absent visual frames or features during evaluation.\footnote{We were able to extract $2,652,463$ frames, which account for $95\%$ of the dataset.}
	
	\textbf{Audio preprocessing.} We sample audio at a conventional $16$kHz rate, and extract $40$-dimensional log-Mel spectrograms. Since the dataset comprises videos recorded at different frame rates ($7.5$ fps to $30$ fps), we adopt the solution proposed in \cite{DBLP:journals/corr/abs-1911-04890} to extract synchronized audio features, by advancing the analysis window at a rate proportional to the video frame rate. By stacking $2$ extra context frames from both directions, we obtain $200$-dimensional feature vectors for each video frame. Videos recorded at $15$ fps or lower are discarded while training with only the audio stream, and a vector of zeros is used in lieu of the features otherwise.
	
	\textbf{Data augmentation.} During training, we take a random $112\times 112$ crop at the same position for each frame, and apply random horizontal flipping to the entire sequence. During evaluation, we take a central $112\times 112$ crop.
	
	\textbf{Evaluation metric.} The official metric for the challenge is the Canonical Concordance Coefficient, which is defined as
	\begin{equation}\label{eq:ccc}
	\rho_c = \frac{2s_{xy}}{s_x^2 + s_y^2 + (\bar x - \bar y)^2},
	\end{equation}
	where $x$ and $y$ are the ground truth annotations and the predicted values, $s_x$ and $s_y$ are their variances, $\bar x$ and $\bar y$ are mean values, and $s_{xy}$ is the covariance. CCC takes values in $[-1, 1]$, where $+1$ indicates perfect concordance and $-1$ indicates perfect discordance. Higher mean valence and arousal CCC is desired for the valence-arousal estimation task.
	\subsection{Experimental Settings}
	We implemented our network in PyTorch~\cite{DBLP:conf/nips/PaszkeGMLBCKLGA19}. The network is trained on servers with NVIDIA Titan Xp GPUs, each with $12$GB memory, and optimized with the Adam optimizer~\cite{DBLP:journals/corr/KingmaB14} using default parameters. The inputs are batches of $64$ clips, normalized to $[-1, 1]$. We use cyclical learning rates~\cite{DBLP:conf/wacv/Smith17} with $\texttt{base\_lr}=10^{-7}$ and $\texttt{num\_step\_up}=3\cdot\texttt{iters\_per\_epoch}$ for single-stream training, and manual LR decay starting with $10^{-5}$ for joint training. Weight decay is set to $0.0001$.
	
	We sample $200$ $32$-frame windows (i.e. $T=32$ in Fig.~\ref{fig:framework}) from each video in the training set during one epoch. During inference, we segment test videos into non-overlapping clips. The model is trained in three stages: first, we train single-modality models with CCC loss (for the audio subnetwork), or with the multi-task loss (for the visual subnetwork). Next, we initialize the fusion network by training for three epochs while keeping the visual and audio encoders frozen. Finally, the network is fine-tuned end-to-end.
	\subsection{Results and Discussion}\label{ssec:results}
	\begin{table}[!t]
		\renewcommand{\arraystretch}{1.3}
		\caption{Results for valence-arousal estimation on the validation set of ABAW Challenge 2020.}
		\label{table:results}
		\centering
		\begin{tabular}{l|c|c|c|c|c}
			\hline
			\multirow{2}{*}{\textbf{Method}} & \multicolumn{2}{c|}{\textbf{Modality}} & \multicolumn{3}{c}{\textbf{CCC~$\uparrow$}}                   \\ \cline{2-6} 
			& \textbf{~V}     & \textbf{A}    & \textbf{Valence} & \textbf{Arousal} & \textbf{Mean} \\ \hline
			Baseline (PatchGAN)~\cite{kollias2020analysing}	& \cmark & \xmark &   $0.14$    &     $0.24$   &    $0.19$    \\ \hline
			AEG-CD-ZS~\cite{asp2020adversarialbased}	& \cmark & \xmark &   $0.10$    &     $0.26$   &    $0.18$    \\ \hline
			SENet-50, fine-tuned~\cite{zhang2020facial}	& \cmark & \xmark &   $0.28$    &     $0.34$   &    $0.31$    \\ \hline\hline
			\multicolumn{6}{l}{\textbf{Audio}}\\\hline
			GRU, scratch & \xmark & \cmark & $0.19$ & $0.36$ & $0.28$ \\
			\hline\hline
			\multicolumn{6}{l}{\textbf{Visual}}\\\hline
			Ours, scratch, w/o MTL & \cmark & \xmark &  &  &  \\ 
			+ SE (V) / TCAE (A) & \cmark & \xmark & $0.27$ & $0.44$ & $0.35$ \\
			+ SE (V) / SE (A) & \cmark & \xmark & $0.26$ & $0.48$ & $0.37$ \\
			~ + VoxCeleb2 pretrain & \cmark & \xmark & $0.36$ & $0.48$ & $0.42$ \\
			\hline
			Ours, scratch & \cmark & \xmark &  &  &  \\ 
			+ SE (V) / SE (A) & \cmark & \xmark & $0.33$ & $0.48$ & $0.40$ \\
			~ + VoxCeleb2 pretrain & \cmark & \xmark & $0.33$ & $0.51$ & $0.42$ \\
			\hline\hline
			\multicolumn{6}{l}{\textbf{Audio-visual} ($M^3$T)}\\\hline
			\textbf{Concat fusion} & \cmark & \cmark & $\mathbf{0.32}$ & $\mathbf{0.55}$ & $\mathbf{0.44}$ \\ \hline
			Attn. fusion & \cmark  & \cmark  &   $0.33$  &  $0.51$   & $0.42$ \\\hline
		\end{tabular}
	\end{table}
	We report our results on the official validation set of the ABAW 2020 Challenge~\cite{kollias2020analysing} in Table \ref{table:results}. Our best performing model achieves a mean CCC of $0.44$, which is a significant improvement over previous results.
	
	We make several observations: first, the proposed model achieves strong arousal estimation performance, which can be attributed to the use of 3D convolutions, which captures temporal dynamics at an early stage; second, similar to the finding in \cite{DBLP:journals/corr/abs-1910-04855}, the combination of audio and video yields noticeable improvements for arousal estimation, but not valence estimation; third, interestingly, better results can be obtained by concatenating SENet-101 features instead of the AU features to the arousal branch. Finally, the attention-based slightly underperforms concatenation fusion. This might be because our models did not train long enough before this submission (we have not reached full model convergence at the time of submission).
	
	\textbf{How does pretraining affect performance?} State-of-the-art results for valence-arousal estimation with static frames use pretrained VGG-Face descriptors~\cite{DBLP:journals/corr/abs-1910-04855,DBLP:journals/ijcv/KolliasTNPZSKZ19}. To the best of our knowledge, we are the first to apply a 3D ConvNet to \textit{in-the-wild} valence-arousal estimation. To understand the role of pretraining in our case, we report results of both training from scratch on Aff-Wild2, and using external pretraining (on VoxCeleb2). As illustrated in Table~\ref{table:results}, similar to the case with 2D ConvNets for VA estimation, pretraining also boosts the performance of our 3D backbone. It can be argued that this improvement is due to the model being initialized with a strong facial feature extractor that is more robust to identity and lighting conditions, which also leads to much faster convergence.
	
	Due to time constraints, the numbers reported for the audio-visual models here do not use VoxCeleb2 pretraining. We will update the corresponding numbers as soon as the results are available.

	\section{Conclusion}
	In this report, we have described a multi-modal, multi-task learning framework named $M^3$T for continuous valence-arousal estimation, which has been used in our entry to the ABAW challenge at FG 2020. The proposed framework leverages 3D and 2D ConvNet visual features, categorical emotion labels, as well as audio information. Our results show significant improvements over the baseline on the ABAW Challenge validation set.
	
	\section*{Acknowledgment}
	We thank Xuran Sun for providing us with the pre-trained SENet FER model.
	
	
	
	\bibliographystyle{IEEEtran}
	\bibliography{bibs}
	
\end{document}